\documentclass[german,english]{bvm2018}

\title{Frangi-Net: A Neural Network Approach to Vessel Segmentation}

\author{Weilin Fu$^1$, Katharina Breininger$^1$, Tobias W\"urfl$^1$, Nishant Ravikumar$^1$, Roman Schaffert $^1$, Andreas Maier$^{1,2}$}

\authorrunning{Weilin Fu et al.}

\institute{%
$^1$Pattern Recognition Lab, Department of Computer Science, Friedrich-Alexander-Universit\"at Erlangen-N\"urnberg \\
$^2$Erlangen Graduate School in Advanced Optical Technologies (SAOT), Erlangen, Germany}

\email{weilin.fu@fau.de}

\begin{document}

%
\selectlanguage{english}

\maketitle

\begin{abstract}
In this paper, we reformulate the conventional 2-D Frangi vesselness measure into a pre-weighted neural network (``Frangi-Net"), and illustrate that the Frangi-Net is equivalent to the original Frangi filter. Furthermore, we show that, as a neural network, Frangi-Net is trainable. We evaluate the proposed method on a set of 45 high resolution fundus images. After fine-tuning, we observe both qualitative and quantitative improvements in the segmentation quality compared to the original Frangi measure, with an increase up to $17\%$ in F1 score.



\end{abstract}

\section{Introduction}


Fundus imaging can help to diagnose and monitor a number of diseases, such as diabetic retinopathy, glaucoma, age-related macular degeneration~\cite{maji2016ensemble}. Visual analysis by a trained ophtamologist can be extremely time-consuming and hinder broad clinical application. To support this workflow, automatic segmentation~\cite{kirbas2004review} of the retinal vessel tree has been studies for decades, among which a vessel enhancement filter by Frangi et al.~\cite{frangi1998multiscale} is the most popular and forms the basis to various other strategies~\cite{budai2013robust}. However, this task is particularly challenging due to the complex structure of retinal vessels (e.g. branching, crossing) and low image quality (e.g. noise, artifacts, low resolution). In recent years, deep learning techniques have been exploited in the field of retinal vessel segmentation~\cite{maji2015deep,tetteh2017deep}. These approaches are data-driven, and tend to share a similar structure, where a classifier follows a feature extractor. For instance, Maji et al.~\cite{maji2015deep} use an auto-encoder as a feature extractor, and random forests as a classifier; Tetteh et al.~\cite{tetteh2017deep} apply a convolutional neural network (CNN) inception model for feature extraction and another CNN model for classification. These novel, data-driven approaches perform comparably to conventional methods, but do not yield significant improvements.

In this paper, we also investigate deep learning techniques on retinal vessel segmentation. To avoid an explicit separation into ``feature extractor" and ``classification", we propose an alternative approach, which formulates the Frangi filter as a pre-weighted network (``Frangi-Net"). The intuitive reasoning behind such an approach is that, by representing the Frangi filter as a pre-weighted neural network, subsequent training of the latter should improve segmentation quality. We aim to utilize prior knowledge of tube segmentation as a basis, and further improve it using a data-driven approach.



\section{Materials and Methods}
%
%

%






\subsection{Frangi Vessel Enhancement Filter}
The Frangi vesselness filter~\cite{frangi1998multiscale} is based on the eigen-value analysis of the Hessian matrix in multiple Gaussian scales. 
Coarse scale structures are typically obtained by smoothing the image with a Gaussian filter $g_\sigma$ where $\sigma$ is the standard deviation. The Hessian matrix is a square matrix of second-order partial derivatives of the smoothed image. Therefore, it can be alternatively calculated by convolving the original image patch directly with a 2-D kernel $G_\sigma$, which is the second-order partial derivatives of the Gaussian filter $g_\sigma$, 
\begin{equation}
\label{eq:hessian}
G_\sigma = \begin{bmatrix}
\frac{\partial ^2 g_\sigma}{\partial x^2} & \frac{\partial ^2 g_\sigma}{\partial x \partial y} \\ 
\frac{\partial ^2 g_\sigma}{\partial x \partial y} & \frac{\partial ^2 g_\sigma}{\partial y^2}
\end{bmatrix}
\end{equation}
The Hessian matrix is calculated as, 
\begin{equation}
H_\sigma = G_\sigma \ast f = \begin{bmatrix}
H_{xx} & H_{xy}\\ 
H_{xy} & H_{yy}
\end{bmatrix}
\end{equation}
The two eigenvalues of the Hessian matrix are denoted by $\lambda_1$ and $\lambda_2$ $(|\lambda_2|\ge|\lambda_1|)$, which are calculated as \begin{equation}\label{eq:lambda}
\lambda_{1, 2} = \frac{(H_{xx}+H_{yy})\pm \sqrt{(H_{xx} - H_{yy})^2+4H_{xy}^2}}{2}
\end{equation}
A high vesselness response is obtained if $\lambda_1$ and $\lambda_2$  satisfy the following conditions: $\|\lambda_1\|\approx 0$, and $\|\lambda_2\|\gg \|\lambda_1\|$. A mathematical description of the vesselness response is presented in~\cite{frangi1998multiscale}, \begin{equation} \label{eqn:Frangi}
V_0(\sigma) = \left\{\begin{array}{ll}
0, & \text{ for dark tubes if } \lambda_2 < 0,\\ 
\exp(-\frac{R_B^{2}}{2\beta ^{2}})(1-\exp(-\frac{S^2}{2c^2})), & \text{ otherwise,}
\end{array}
\right.
\end{equation} where $S = \sqrt{\lambda_1^2 + \lambda_2^2}$ is the second order structureness, $R_B = \frac{\|\lambda_1\|}{\|\lambda_2\|}$ is the blobness measure, and $V_0$ stands for the vesselness value. $\beta, c$ are image-dependent parameters for blobness and structureness terms, and are set to 0.5, 1, respectively.

When using a Gaussian kernel with a standard deviation $\sigma$, vessels whose diameters equal to $2\sqrt{2}\sigma$ have the highest vesselness response. The diameter of retinal vessels in this work range from 6 to 35 pixels. Hence we choose a series of $\sigma$ as $3, 6, 12$ pixels accordingly. Instead of convolving a patch with three different kernels (i.\,e., $G_3, G_6, \text{and}~G_{12}$), we create a three-level resolution hierarchy~\cite{budai2013robust}, and convolve each level with $G_3$ only. The resolution hierarchy consists of the original patch and two downsampled versions, using factors 2 and 4, respectively.

\subsection{Pre-weighted Frangi-Net}
%


Inspired by~\cite{wurfl2016deep}, we implement the multi-scale Frangi filter as a neural network called Frangi-Net on the basis of the previous section. The architecture of Frangi-Net is described in Fig.~1. The proposed method is used to analyze image patches, with an output size set to $128 \times 128$ pixels. Input patches for the three sub-nets are cropped as in Fig.~1 (a).




For a single-scale sub-net, we first use a resize layer to downsample images to the corresponding size. Secondly, we use a 2-D convolution layer with three filters to get the Hessian matrix. These filters are initialized as second partial derivatives of the two-dimensional Gaussian function with $\sigma=3$. After that, a combination of mathematic operation layers are applied, based on Eqs.~\ref{eq:lambda} and~\ref{eqn:Frangi} to calculate eigen-values and vesselness responses of this scale.


\begin{figure}[htb]
\label{frameworks}
	\setlength{\figbreite}{0.42\textwidth}
	\centering
	\subfigure[Patch Set]{\includegraphics[width=\figbreite]{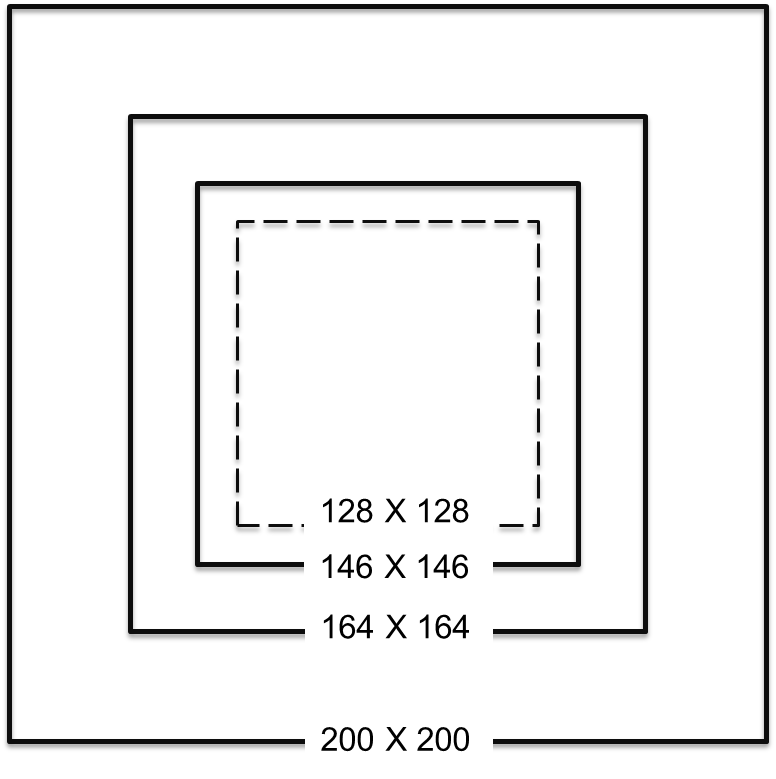}}
	\subfigure[Single-scale Net Structure]{\includegraphics[width=\figbreite]{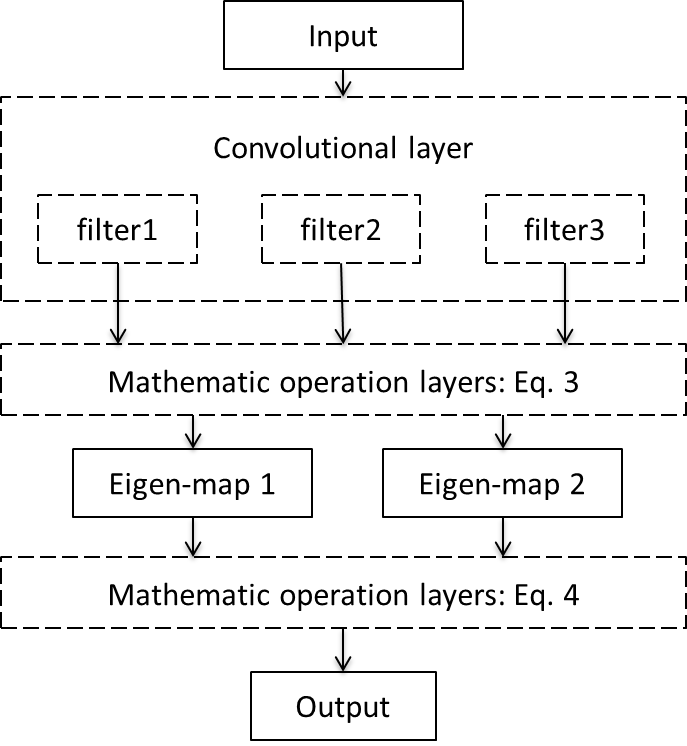}}
	\subfigure[Frangi-Net Overall Structure]{\includegraphics[height=\figbreite]{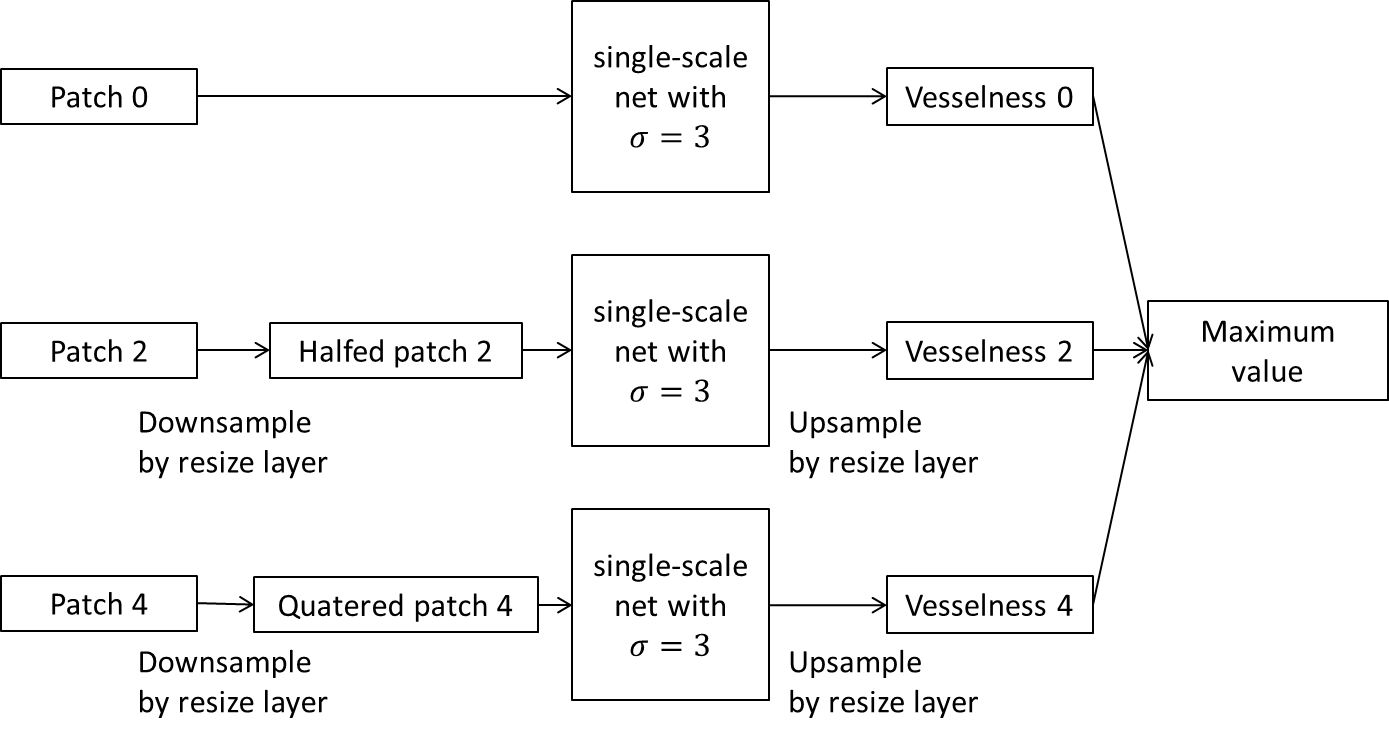}}
	\caption{The Frangi-Net architecture: In (a), dash line represents output patch size $128\times128$; solid lines represent the patch set in the resolution hierarchy. (b) describes the dataflow through the framework of a single-scale net, where eigen-map 1 and 2 are maps of $\lambda_1, \lambda_2$ respectively; (c) describes the overall structure of Frangi-Net, where patch 0, patch 2, patch 4 correspond to size $146\times146$, $164\times164$, $200\times200$ in (a).}	
\end{figure} 

The raw vesselness scores from Frangi-Net typically range from 0 to roughly 0.4. To create a binary vessel mask, a threshold $t$ is used, which is set to $10^{-3}$ in this work. This means that most background pixels are squashed inbetween $[0, t)$, and the rest few vessel pixels spread along $(t, 0.4)$. To obtain a probability map, we subtract the data with the threshold $t$, and asymmetrically scale it, where negative values are multiplied with $20,000$ and positive values with $2000$ so that raw scores are redistributed to $[-20, 80)$. Afterwards, we apply a sigmoid layer. In this way, raw vesselness values of $0, t,$ and $0.4$ are converted to $0, 0.5,$ and 1, respectively. 

\subsection{Trainable Frangi-Net}

With the proposed network structure, we observe two sets of trainable parameters in the network: first, the covolution kernels responsible for the computation of the Hessian matrix; second, the parameters $\beta, c$, that control the influence of structureness and blobness features. For each scale, this results in three trainable convolution kernels and two additional parameters that can be adapted during training. In Frangi-Net, single-scale nets are structured in parallel and updated independently, which means in total we have nine kernels and six parameters to update during training.




 
A high resolution fundus (HRF) image database~\cite{dubai} is used in this work. The database includes 15 images of each healthy, diabetic retinopathy, and glaucomatous subjects. The HRF images are high resolution $3504 \times 2336$ pixel RGB fundus photographs. Only the green channels are used where vascular structures manifest a good contrast~\cite{yin2015vessel}. The intensity values are normalized to [0, 1]. We train Frangi-Net on each cartegory independently, randomly dividing the 15 datasets into 10 training, 2 validation and 3 testing sets. We also train the network with all 45 datasets, randomly taking 30 sets for training, 6 for validation and 9 for testing. As segmentation of retinal vessels from fundus images is an unbalanced classification problem, where the background pixels outnumber vessel pixels by approximately 10 to 1, dice coefficient loss~\cite{taha2015metrics} instead of the common cross entropy loss is used. The optimizer we utilize here is gradient descent with learning rate $10^{-6}$ and momentum 0.5. 1000 steps are used for training on each category independently, while 3000 steps are used for training on the whole database. Batch size is chosen as 250. We use python and tensorflow framework for the whole implementation.


\section{Results}

A quantitative comparison of the trained Frangi-Nets with the original Frangi filter is summarized in Table~\ref{resulttable}. Each quality metric is calculated as an average of that of the testing datasets.
After fine-tuning with four different training datasets, the F1 score, accuracy, precision and recall of the segmentation results on the testing datasets all get improved. Segmentation of heathy fundus images performs best, achieving a highest post-training F1 score as 0.712. Comparing the diseased cases to the healthy cases, the F1 score has a more significant improvement after training, up to $17\%$. 

\begin{figure}[htb]
	\setlength{\figbreite}{0.45\textwidth}
	\centering
	\subfigure[Before]{\includegraphics[width=\figbreite]{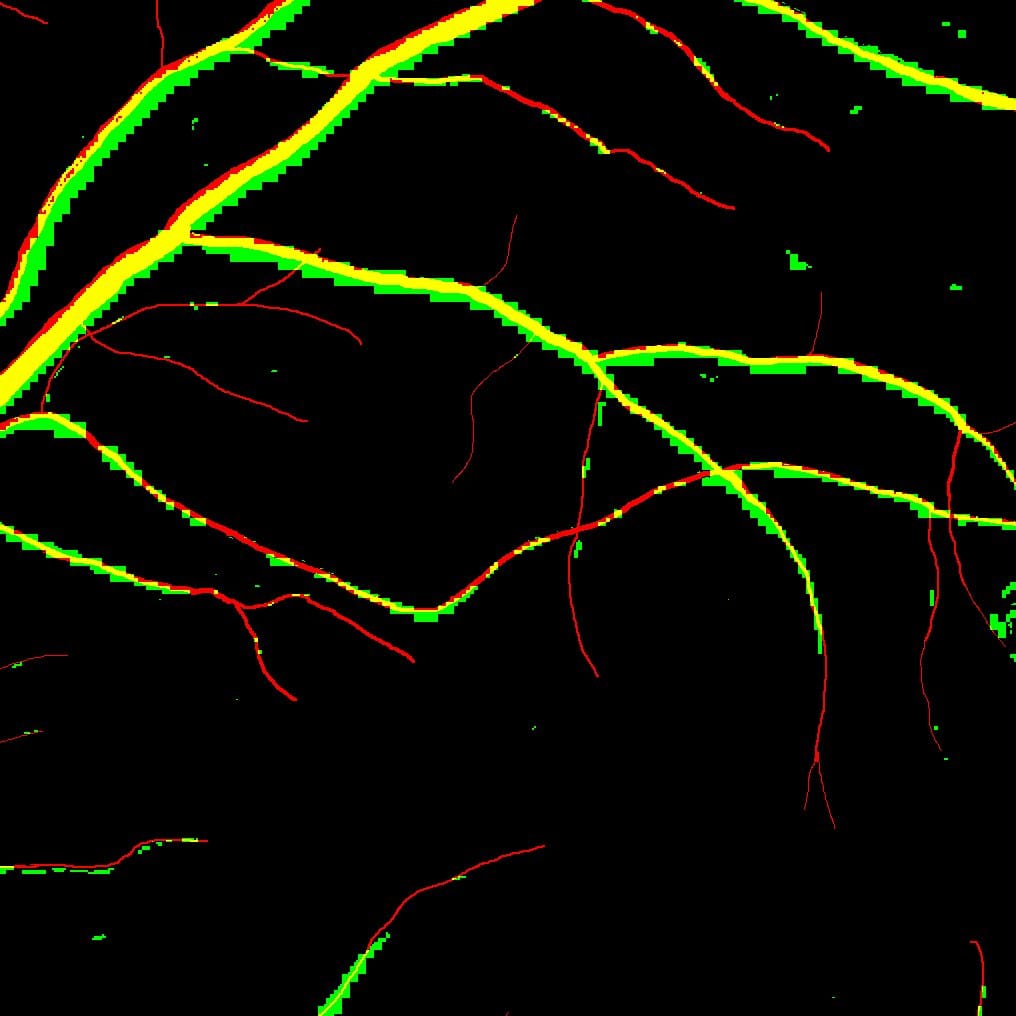}}
	\subfigure[After]{\includegraphics[width=\figbreite]{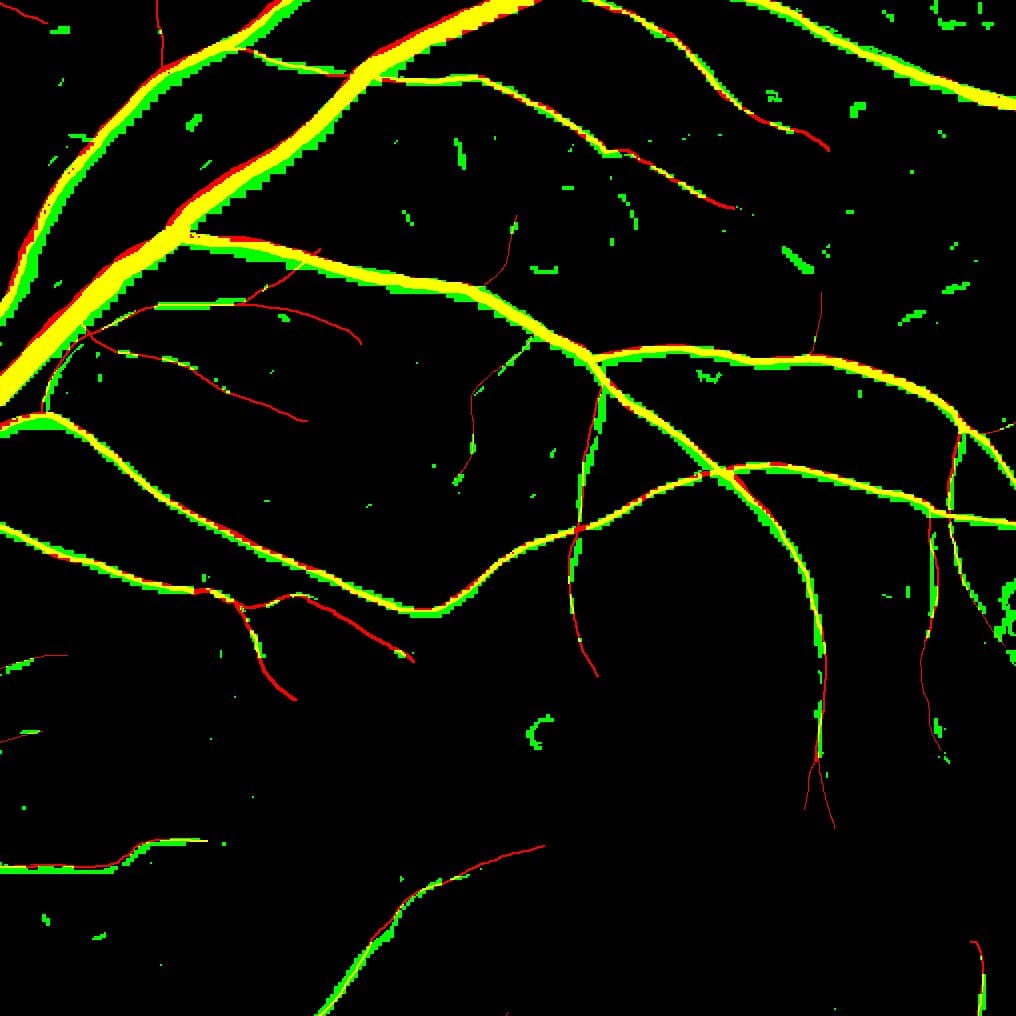}}
	\caption{Vessel segmentation results before and after training. The red, green and yellow colors represent the manual labels, the segmentation results and the overlaps between them, respectively.}\label{img_prepost}
\end{figure}

A comparison of the segmentation results before and after training using the healthy datasets is shown in~Fig.\ref{img_prepost}. Fig.~\ref{img_prepost}(a) displays that Frangi-Net can segment main vessels well before training. However, it only partially segments the middle-sized vessels and fails to recognize most of the tiny ones. In contrast, after training, Frangi-Net is able to detect the middle-sized vessels well, and it also detects most of the tiny ones (Fig.~\ref{img_prepost}(b)).



\begin{table}[]
\centering
\caption{Segmentation quality before and after training}
\label{resulttable}
\begin{tabular}{|l|c|c|c|c|c|c|c|c|}
\hline
\multirow{2}{*}{Dataset} & \multicolumn{2}{c|}{F1 score}                            & \multicolumn{2}{c|}{accuracy}                            & \multicolumn{2}{c|}{precision}                           & \multicolumn{2}{c|}{recall}                              \\ \cline{2-9} 
                         & \multicolumn{1}{l|}{before} & \multicolumn{1}{l|}{after} & \multicolumn{1}{l|}{before} & \multicolumn{1}{l|}{after} & \multicolumn{1}{l|}{before} & \multicolumn{1}{l|}{after} & \multicolumn{1}{l|}{before} & \multicolumn{1}{l|}{after} \\ \hline
Healthy                  & 0.669                      & \textbf{0.712}            & 0.836                      & \textbf{0.843}            & 0.606                      & \textbf{0.675}            & 0.746                      & \textbf{0.753}            \\ \hline
Diabetic retinopathy     & 0.495                      & \textbf{0.532}            & 0.819                      & \textbf{0.822}            & 0.524                      & \textbf{0.588}            & 0.468                      & \textbf{0.486}            \\ \hline
Glaucomatous             & 0.495                      & \textbf{0.584}            & 0.838                      & \textbf{0.841}            & 0.477                      & \textbf{0.562}            & 0.608             & \textbf{0.618}\\ \hline
Whole dataset            & 0.612                      & \textbf{0.672}            & 0.847                      & \textbf{0.855}            & 0.603                      & \textbf{0.675}            & 0.623                      & \textbf{0.670}            \\ \hline
\end{tabular}
\end{table}

\section{Discussion}
\label{formatvorlagen}

In this work, we proposed to combine the prior knowlegde about retinal vessel that is encoded in the Frangi-Filter with the data-driven capabilities of neural networks. We constructed a net that is equivalent to the multi-scale Frangi filter. We identified the trainable parts of Frangi-Net as convolutional kernels and parameters when computing vesselness. We redistributed the raw output vesselness values into a probability map and trained the net by optimizing dice coefficient loss. In experiments with high resolution fundus images of healthy and diseased patients, the trained Frangi-Net performs better than the original formulation both quantitatively and qualitatively. 



Future work will investigate on different network architectures for Frangi-Net, or the combination of Frangi-Net with other networks. We will also evaluate Frangi-Net with other datasets.


\bibliographystyle{bvm2018}

\bibliography{0000}
\end{document}